\documentclass[12pt]{article}
\usepackage{geometry}

\newgeometry{vmargin={15mm}, hmargin={22mm,30mm}}
\usepackage{graphicx}
\usepackage{multirow}
\usepackage{hyperref}

\usepackage{graphicx}
\usepackage{amsthm}

\newtheorem{theorem}{Theorem}
\newtheorem{remark}{Remark}

\usepackage{amsmath}
\usepackage{amssymb}
\usepackage{epsfig}
\usepackage{float}
\usepackage{color}
\usepackage{booktabs}
\providecommand{\keywords}[1]{\textbf{\textit{Index terms---}} #1}

\newcommand{\drop}[1]{}

\newcommand{\fer}[1]{(\ref{#1})}
\newcommand{\qtext}[1]{\quad\text{#1}}

\newcommand{\eps}{\varepsilon}

\newcommand{\grad}{\nabla}

\newcommand{\nl}{\text{\emph{nl}}}

\newcommand{\R}{\mathbb{R}}

\def\O{\Omega}

\newcommand{\abs}[1]{| #1 |}

 \DeclareMathOperator{\Div}{div}

\usepackage{footmisc}
\DefineFNsymbols{mySymbols}{{\ensuremath\dagger}{\ensuremath\ddagger}\S\P
   *{**}{\ensuremath{\dagger\dagger}}{\ensuremath{\ddagger\ddagger}}}
\setfnsymbol{mySymbols}
\begin{document}






\title{Non-convex non-local flows for saliency detection}
\author{I. Ram\'{\i}rez\thanks{Dpt. of Mathematics, Universidad Rey Juan Carlos,
  28993-Madrid, Spain} \\  \href{mailto:ivan.ramirez@urjc.es}{\small ivan.ramirez@urjc.es}  \and  G. Galiano\thanks{Dpt. of Mathematics, Universidad de Oviedo,
33007-Oviedo, Spain} \\  \href{mailto:galiano@uniovi.es}{\small galiano@uniovi.es} \and  E. Schiavi\footnotemark[1] \\  \href{mailto:emanuele.schiavi@urjc.es}{\small emanuele.schiavi@urjc.es}}




\maketitle
\begin{abstract}
We propose and numerically solve a new variational model for automatic saliency detection in digital images. Using a non-local framework we consider a family of edge preserving functions combined with a new quadratic saliency detection term. Such term defines a constrained bilateral obstacle problem for image classification driven by $p$-Laplacian operators, including the so-called hyper-Laplacian case ($0<p<1$). The related non-convex non-local reactive flows are then considered and applied for glioblastoma segmentation in magnetic resonance fluid-attenuated inversion recovery (MRI-Flair) images. A fast convolutional kernel based approximated solution is computed. The numerical experiments show how the non-convexity related to the hyper-Laplacian operators provides monotonically better results in terms of the standard metrics.
\vspace{0.5cm}

\noindent \keywords{Variational Methods, Non-local Image Processing, Automatic Saliency Detection, MRI-Flair Glioblastoma Segmentation. }
\end{abstract}
\section{Introduction}
\label{sec:introduction}
Image processing by variational methods is an established field in applied mathematics and computer vision which aims to model typical low and mid level image reconstruction and restoration processes such as denoising, deconvolution, impainting, segmentation and registration of digital degraded images. 
Since the mathematical approach of Tikhonov and Arsenin on ill-posed inverse problems \cite{bell1978solutions} and the applied work of Rudin, Osher and Fatemi \cite{rudin1992nonlinear} through their celebrated ROF model, there have been a standing effort to deal with new image processing tasks through variational methods. 
\noindent Currently there is a growing interest in image processing and computer vision applications for visual saliency models, able to focus on perceptually relevant information within digital images. 

Semantic segmentation, object detection, object proposals, image clustering, retrieval and cognitive saliency applications, such as image captioning and high-level image understanding, are just a few examples of saliency based models. 
Saliency is also of interest as a means for improving computational efficiency and for increasing robustness in clustering and thresholding. 

Despite of the lack of a general consensus on a proper mathematical definition of saliency, it has a clear biologically perceptive meaning: it models the mechanism of human attention, that is, of finding relevant objects within an image. 


Recently, there has been a burst of research on saliency due to its wide application in leading medical disciplines such as neuroscience and cardiology. In fact, when considering medical images like those acquired in magnetic resonance imaging (MRI) or positron emission tomography (PET), the automatic obtention of saliency maps is useful for pathology detection, disease classification \cite{rueda2013saliency}, location and segmentation of brain strokes, gliomas, myocardium detection for PET images, tumors quantification in FLAIR MRI \cite{thota2016fast}, etc. 

The role of saliency in models is application dependent, and several different techniques and approaches have been introduced to construct saliency maps. They vary from low dimensional manifold features minimization \cite{zhan2011nonlocal} to non-local sparse minimization \cite{wang2014saliency}, graphs techniques \cite{harel2007graph}, partial differential equations (PDE) \cite{li2013nonlocal}, superpixel \cite{liu2013superpixel}, learning \cite{liu2014adaptive}, or neural networks based approaches \cite{mit-saliency-benchmark} (MIT-Benchmark).

With the aim to explore the applications and algorithms of non-smooth, non-local, non-convex optimization of saliency  models, we present in this article a new variational model applied to Fluid Attenuated Inversion Recovery (FLAIR) MR images for accurate location of tumor and edema. 

Models for saliency detection try to transform a given image, $f$, defined in the pixel domain, $\O$, into a constant-wise image, $u$, whose level sets correspond to salient regions of the original image. They are usually formulated through the inter-relation among three energies: fidelity, regularization, and saliency, being the latter the mechanism promoting the classification of pixels into two or more classes. There is a general agreement in considering the fidelity term as determined by the $L^2$ norm, that is 

\begin{align*}
 F(u) = \frac{1}{2}\int_\O \abs{u-f}^2,
\end{align*}
so that departure from the original state is penalized in the minimization procedure.

For regularization, an edge preserving energy should be preferred. The use of the Total Variation energy,
\begin{align*}
 TV(u)=\int_\O \abs{\nabla u},
\end{align*}
defined on the space of Bounded Variation functions has been ubiquitous since its introduction as an edge preserving restoration model \cite{rudin1992nonlinear}. Indeed, the Total Variation energy allows discontinuous functions to be solutions of the corresponding minimization problem, with discontinuities representing edges. This is in contrast to Sobolev norms, which enforce continuity across level lines and thus introduce image blurring.
Only recently, the non-local version of the Total Variation energy and, in general, of the energy associated to the $p$-Laplacian, 
for $p> 1$, has been considered in restoration modeling. In saliency modeling,  only the range $p>2$ seems to have been treated \cite{li2013nonlocal}. One of the main advantages of introducing these non-local energies is the lack of the hard regularizing effect influencing their local counterparts.

For the saliency term, a phase-transition Ginzburg-Landau model consisting in a double-well absorption-reaction term is often found in the literature \cite{ginzburg1955theory}. 
The resulting energy is a functional of the type
\begin{align*}
\int_\O \big(1-\abs{u}^2\big)^2,
\end{align*}
whose minimization drives the solution towards the discrete set of values $\{-1,1\}$, facilitating in this way the labeling process. However, due to the vanishing slope of 
$g(u)=(1-\abs{u}^2)^2$ at $u=\pm 1$, the resulting algorithm has a slow convergence to the minimizer. 

In this article, we introduce new regularizing and saliency terms that enhance the convergence of the classification algorithm while keeping a good quality compromise.
As regularizing term, we propose the non-local  energy associated to the $p$-Laplacian for any $p>0$, see Eq. \fer{nlpl}. Although there is a lack of a sound mathematical theory for the concave range $0<p<1$, computational evidence of the ability of these energies to produce sparse gradient solutions has been shown \cite{krishnan2009fast}. In any case, our analysis also includes the  energies arising for $p\geq1$, enjoying a well established mathematical theory.

With respect to the saliency term, we consider the combination of two effects. One is captured by the concave energy  
\begin{align*}
H(u)= -\frac{1}{2}\int_\O (1-\delta u)^2,
\end{align*}
with $\delta>0$ constant,
which fastly drives the minimization procedure so that $H(u)\to -\infty$. To counteract this tendency and remain in the meaningful interval $u\in I=[0,1]$, being  $\{0,1\}$ the saliency labels employed in this article, we introduce an obstacle which penalizes the minimization when the solution lies outside $I$. The modeling of such obstacle is given in terms of the indicator function  
\begin{align*} 
\mathcal{I}_I(u) = \left \{ \begin{matrix} 0 & \mbox{if }u \in I,
 \\[0.2cm]
\infty & \mbox{if } u \notin I, \end{matrix}\right.
\end{align*}
and the resulting saliency term is then defined as a weighted sum of the operators $H(u)$ and $\mathcal{I}_I(u)$.

Finally, the Euler-Lagrange equation for the addition of these three energies leads to a formulation of a multi-valued non-local reaction-diffusion problem which is later rendered to a single-valued equation by means of the Yosida's approximants. The introduction of a gradient descent and the discretization of the corresponding evolution equation are the final tools we use to produce our saliency detection algorithm.


The main contributions contained in this paper may be summarized as follows. 
\begin{itemize}
\item The non-local $p$-Laplacian convex model proposed in \cite{li2013nonlocal}, valid for $p>2$, is extended to include the range $0<p\leq 2$. Notably, the non-differentiable non-convex fluxes related to the range $0<p< 1$ are investigated.
\item An absorption-reaction term of convex-concave energy nature is considered for fast saliency detection. In absence of a limiting mechanism, the reactive part could drive the solution to take values outside the image range $[0,1]$. Thus, a penalty term is introduced to keep the solution within such range. 
\item The resulting multivalued constrained variational formulation is approximated and shown to be stable.
\item A fast algorithm overcoming the computational drawbacks of non-local methods is presented. 
\item A 3D generalization of the model is considered providing promising results specially valuable in medical image processing and tumor detection.
\end{itemize}

The paper is organized as follows. 
In Section~\ref{sec:framework}, we introduce the variational mathematical framework 
of our models. Starting with the local equations as guide for the modelling exercise, we focus on the non-local diffusive terms, explicited in the form of $p$-Laplacian, for $p>1$ and extend it to the range $0<p \leq 1$ through a differentiable family of fluxes to cover the resulting non-local non-convex hyper-Laplacian operators. 
Then, we introduce a multi-valued concave saliency detection term which defines an obstacle problem for the non-local diffusion models. In Section \ref{model} we deduce the corresponding Euler-Lagrange equations. A gradient descent approximation is finally used to solve the elliptic non-local problems until stabilization of the associated evolution problems. 
In Section~\ref{sec:discretization}, we give the discretization schemes used for the actual computation of the solutions of the non-local diffusion problems. In Section \ref{cinco} a simplified computational approach is described in order to reduce the time execution with a view to a massive implementation on the proposed data-set. Section \ref{sec:experiments} contains the numerical experiments on the proposed model and present the simulations performed on FLAIR sequences of MR images obtained from the BRATS2015 dataset \cite{menze2015multimodal}. 
Finally, in Section~\ref{sec:conclusion}, we give our conclusions.

\section{Variational framework}\label{sec:framework}

\subsection{Local $p$-Laplacian}

Variational methods have shown to be effective to model general low level computer vision processing tasks such as denoising, restoration, registration, deblurring, segmentation or super-resolution among others. A fundamental example is given by the minimization of the energy functional
\begin{equation}
\label{prob:min}
E_p (u)=J_p(u) + \lambda F(u),
\end{equation}
where $\lambda>0$ is a constant, $J_p(u)$ and $F(u)$ are the \emph{regularization} and the \emph{fidelity} terms, respectively, given by
\begin{equation*}
 J_p(u) = \frac{1}{p}\int_\O \abs{\grad u}^p,\quad F(u) = \frac{1}{2} \int_\O \abs{u-f}^2,
\end{equation*}
$\Omega\subset \R^2$ is the set of pixels, $f:\Omega\to[0,1]$ is the image to be processed, and $u:\Omega\to\R$ belongs to a space of functions for which the minimization problem admits a solution. The idea behind this minimization problem is: given a non-smooth (e.g. noisy) image, $f$, obtain another image which is close to the original (fidelity term) but regular (bounded gradient in $L^p(\O)$). The parameter $\lambda$ is a weight balancing the respective importance of the two terms in the functional. 

When first order necessary optimality conditions are imposed on the energy functional, a PDE (the Euler-Lagrange equation) arises. In our example, 
\begin{equation}
\label{eq:el-l}
-\Div\big(\abs{\grad u}^{p-2} \grad u \big) + \lambda (u-f)=0.
\end{equation}
The divergence term in this equation is termed as \emph{$p$-Laplacian}, and its properties have been extensively studied in the last decades for the range of exponents $p\geq 1$. For $p>1$, the energy $J_p(u)$ is convex and differentiable, and the solution to the minimization problem belongs to the Sobolev space $W^{1,p}(\O)$, implying that $u$ can not have discontinuities across level lines. Therefore, the solution, $u$, is smooth  even if the original image, $f$, has steep discontinuities (edges). This effect is known as \emph{blurring}: the edges of the resulting image are diffused.

In the case $p=1$, the energy term $J_1(u)$ is convex but not differentiable. The solution of the related minimization problem (the Gaussian Denoising model \cite{rudin1992nonlinear,chambolle1997image} or the Rician Denoising model \cite{martin20171}) belong to the space of functions of Bounded Variation, $BV(\O)$, among which the constant-wise functions play an important role in image processing tasks  \cite{ambrosio2000functions}. Thus, in this case, the edges of $f$ are preserved in the solution, $u$, because a function of bounded variation may have discontinuities across surface levels.

In this article, we are specially interested in the range $0<p<1$, for which the energy $J_p(u)$ is neither convex nor differentiable, and it only generates a quasi-norm on the corresponding $L^p (\Omega )$ space.  In this parameter range the problem lacks of a sound mathematical theory, although some progress is being carried on  \cite{hintermuller2014smoothing}. 
Despite the difficulties for the mathematical analysis, there is numerical evidence on interesting properties arising from this model. In particular, the non-convexity forces the gradient to be \textit{sparse} in so far it minimizes the number of jumps in the image domain. Actually, only sharp jumps  are preserved, looking the resulting image like a cartoon piecewise constant image.

\subsection{Nonlocal $p$-Laplacian}

While for $p>1$ the use of the local $p$-Laplacian energy is not specially relevant in image processing due to its regularizing effect on  solutions which produces over-smoothing of the spatial structures,  for its non-local version the initial data and the final solution belong to the same functional space, that is, no global regularization takes place. See \cite{andreu2010nonlocal}, where a thorough study on non-local diffusion evolution problems, including  existence and uniqueness theory, may be found.

The non-local analogous to the energy $J_p(u)$, for $p>1$,  is 
\begin{align}\label{nlpl}
J_p^{\nl} (u) =\frac{1}{2p}\int_{\Omega \times \Omega} w(x-y)\displaystyle | u(y)-u(x)|^{p}  dydx, 
\end{align}
where $w$ is a continuous non-negative radial function with $w(0)>0$ and $\int_\O w =1$.
The Fr\'echet differential of $J_p^{\nl} (u)$ is  
\begin{align*}
 DJ_p^{\nl} (u)=\int_\Omega w(x-y)\displaystyle | u(y)-u(x)|^{p-2} (u(y)-u(x)) dy. 
\end{align*}
Thus, the Euler-Lagrange equation for the minimization problem \fer{prob:min} when $J_p(u)$ is replaced by $J_p^{\nl} (u)$ is  
\begin{equation}
\label{eq:el-nl}
 \int_\Omega w(x-y)\displaystyle | u(y)-u(x)|^{p-2} (u(y)-u(x)) dy + \lambda (f-u)=0.
\end{equation}


For $p\leq 1$, the Euler-Lagrange equation  \fer{eq:el-nl}  does not have a precise meaning due to the singularities that may arise when the denominator vanishes. To overcome this situation, we approximate the non-differentiable energy functional  $J_p^{\nl} (u)$ by 
\begin{align*}
 J_{\epsilon ,p}^{\nl} (u) & =\frac{1}{4} \int_{\Omega \times \Omega} w(x-y)\phi_{\epsilon ,p} (u(y)-u(x))dxdy,
\end{align*}
for $\epsilon >0$, where 
\begin{equation*} 
\phi_{\epsilon ,p} (s) = \displaystyle \frac{2}{p}\displaystyle \left( s^2 +\epsilon^2\right)^{p/2} -\frac{2}{p}\epsilon^p .
\end{equation*}
Observe that the corresponding minimization problem is now well-posed due to the differentiability of $J_{\epsilon ,p}^{\nl} (u)$. Therefore, a solution may be calculated solving the associated Euler-Lagrange equations. However, for $p<1$, the solution is in general just a local minimum, due to the lack of convexity. Of course, the same plan may be followed for the local diffusion equation \fer{eq:el-l}.

\subsection{Saliency modeling}\label{sec:saliency}
The previous section highlighted the connections between local and non-local formulations. From now on, we focus on the non-local diffusion model, being the model deduction similar for the case of local diffusion. 


For saliency detection and classification, an additional term is added to the energy functional \fer{prob:min} or to its non-local or regularized variants. The general idea is  
pushing the values of $u$ towards the discrete set of extremal image values $\{0,1\}$, 
determining 
the labels we impose for saliency detection: $u = 1$ for foreground, and $u = 0$ for background.

To model this behavior we propose a two-terms based energy, where the first causes a reaction extremizing the values of the solution and the second accounts for the problem constraints ($0\leq u\leq 1$). 
The first term is given by the energy
\begin{equation*}
H(u) =  \int_\Omega h(u),\quad\text{with}\quad h(u)=-\frac{1}{2}(1-\delta u)^2,
\end{equation*}
for some constant $\delta >0$.
The corresponding energy minimization drives the solution away from the zero maximum value, attained at $u= 1/\delta$,  and it would be unbounded ($-\infty$) if no box constraint were assumed. 

However, under the box constraint, the global minimum of $h(u)$ in $[0,1]$ is attained at the boundary of this interval, that is, at the labels of saliency identification. 
Therefore, $H(u)$ together with the box constraint promote the detection of salient regions of interest (foreground) separated by regions with no relevant information (background).


The box constraint is accomplished through the indicator functional  $\mathcal{I}_I(u)$ of the interval $I=[0,1]$, defined as

\[ 
\mathcal{I}_I(u) = \left \{ \begin{matrix} 0 & \mbox{if }u \in I,
	\\
\infty & \mbox{if } u \notin I. \end{matrix}\right. 
\]
Observe that since $I$ is convex and closed, the functional $\mathcal{I}_I(u)$ is convex and lower semi-continuous, and that its sub-differential is the maximal monotone graph of $\mathbb{R}\times \mathbb{R}$, given by  
\begin{equation*}
\partial \mathcal{I}_I(u)=
\begin{cases}
 (-\infty ,0] & \mbox{if }u =0,\\
0 & \mbox{if } 0<u<1 ,\\
[0,+\infty ) & \mbox{if } u=1, \\
\emptyset & \mbox{otherwise.}
\end{cases}
\end{equation*}

The saliency term we propose is therefore the sum of the fast saliency promotion, $H(u)$, and of the range limiting mechanism, ${\cal I}_I (u)$, i.e.   
\begin{align*}
S(u)=H (u)+{\cal I}_I (u).
\end{align*}

\section{The model. Approximation and estability}\label{model}

Gathering the fidelity, the regularizing and the saliency energies, we define a bilateral constrained obstacle problem associated to the
following energy
\begin{align}
\label{def:e-nl}
 E_{\epsilon,p}^{nl} (u)=\alpha J_{\epsilon,p}^{nl}(u)+\lambda F(u) + \frac{1}{\alpha}S(u),
\end{align}
where $\alpha>0$ is a parameter modulating the relationship between regularization and saliency promotion. Observe that there is no use in multiplying $\mathcal{I}_I(u)$ by the constant $1/\alpha$, so we omit it for clarity.

The Euler-Lagrange equation corresponding to \fer{def:e-nl} together with the use of a gradient descent method leads to the consideration of non-local multi-valued evolution problems. 

\vspace{0.2cm}
\noindent \textbf{Multi-valued Problems $P_\epsilon(u)$}
\vspace{0.2cm}

\noindent Let $\alpha$, $\delta$, $\lambda$ and $\epsilon$ be real fixed positive parameters. Let moreover $f \in L^{\infty}(\Omega)$ be an (essentially) bounded function. For some given $T>0$, find $u:[0,T]\times \Omega\to\mathbb{R}$ solving the approximating smooth (in fact differentiable) multivalued problems  
\begin{align}\label{discret_model}
P_\epsilon(u) =
\begin{cases}
\partial_tu -  \alpha K_{\eps,p}(u) +\partial \mathcal{I}_I(u) \ni au -b, \qtext{in }(0,T)\times\Omega. \vspace{0.2cm} \\ 
u(0,\cdot) = f \qtext{ on } \Omega
\end{cases}
\end{align}
which model non-linear non-local non-convex reactive flows that we shall consider in the range $0<p\leq 1$.

For the sake of presentation, we have introduced the following notation in \fer{discret_model}: we rewrote the Fr\'echet differential of $H(u)/\alpha +\lambda F(u)$ as $au(x)-b(x)$, with 
\begin{equation}
\label{def:ab}
 a = \frac{\delta^2 }{\alpha}-\lambda  ,\quad b(x)= \frac{\delta }{\alpha}-\lambda f(x), \qtext{for }x\in\O,
\end{equation}
and defined the non-local hyper-Laplacian ($0<p<1$) and $1$-Laplacian ($p=1$) diffusion operators 
\begin{align*}
K_{\epsilon,p}(u)(t,x)= \int_\Omega w(x-y)k_{\epsilon ,p} (u(t,y)-u(t,x))dy,
\end{align*}
with differential kernels
\begin{equation}\label{ker}
 k_{\epsilon ,p} (s)=\frac{1}{2}\phi_{\epsilon ,p}' (s)=\displaystyle s\left(s^2 +\epsilon^2\right)^{\frac{p-2}{2}}.
\end{equation}
Notice that, while for the local diffusion problem we must explicitly impose the homogeneous Neumann boundary conditions, which are the most common boundary conditions for image processing tasks, for the non-local diffusion problem this is no longer necessary since    these conditions are implicitly imposed by the non-local diffusion operator \cite{andreu2010nonlocal}.

\subsection{Yosida's approximants }
Introducing the maximal monotone graphs $\beta,\gamma \subset \R\times\R$ given by
\begin{align*}
 \beta(u)=
  \begin{cases}
   \emptyset & \text{if }u<0,\\
   (-\infty,0] & \text{if } u=0,\\
   0 & \text{if }u>0,
  \end{cases}
  \qquad
 \gamma(u)=
 \begin{cases}
   0 & \text{if }u<0,\\
   [0, \infty) & \text{if } u=0,\\
   \emptyset & \text{if }u>0,
  \end{cases}
\end{align*}
we may express the subdifferential of $\mathcal{I}_I(u)$ as 
$\partial \mathcal{I}_I(u) = \beta(u)+\gamma(u-1)$. 
The Yosida's approximants of $\beta$ and $\gamma$ are then   
\begin{align*}
 \beta_r(u)=
  \begin{cases}
   u/r & \text{if }u\leq0,\\
   0 & \text{if }u>0,
  \end{cases}
  \qquad
 \gamma_r(u)=
 \begin{cases}
   0 & \text{if }u<0,\\
   u/r & \text{if }u\geq0,
  \end{cases}
\end{align*}
for $r>0$,  allowing us to approximate the multi-valued formulation \fer{discret_model} by single-valued equations in which $\beta$ and $\gamma$ are replaced by $\beta_r$ and $\gamma_r$. This is, by the evolution PDE
\begin{equation}\label{eq:yosida}
\partial_tu -  \alpha K_{\eps,p}(u) +\beta_r(u)+\gamma_r(u-1) = au -b .
\end{equation} 
Assume that a solution, $u$, of \fer{eq:yosida} with initial data $u(0,\cdot)=f$ does exist, and 
consider the characteristic function of a set $C$, defined as $\chi_C(x) = 1$ if $x\in C$ and $\chi_C(x)=0$ otherwise. Introducing the sets 
\begin{align*}
\Omega_0(t) =\{ x\in\Omega \,|\, u (t,x)>0\,\},\qquad \Omega_1(t) =\{ x\in\Omega \,|\, u (t,x)<1\,\},
\end{align*}
for  $t\in[0,T)$, we may express the Yosida's approximants appearing in \fer{eq:yosida} as 
\begin{align*}
\beta_r(u(t,x)) = \frac{1}{r}u(t,x)\chi_0(t,x)\quad
\gamma_r(u(t,x)) = \frac{1}{r}(u(t,x)-1)\chi_1(t,x),
\end{align*}
with, for $i=0,1$, $\chi_i(t,x) = \chi_{\Omega\setminus\Omega_i(t)}(x)$. Then, we rewrite \fer{eq:yosida} as a family of approximating problems $P_{\epsilon,r}(u)$.

\vspace{0.2cm}
\noindent \textbf{Approximating Single-Valued Problems $P_{\epsilon,r}(u)$}
\vspace{0.2cm}

\noindent Set $\mathcal{Q}_T$ = $(0,T)\times\Omega$. Given $a>0$, $f \in L^\infty (\Omega)$, $f(x)\geq 0$ a.e. in $\Omega$, define $b(x) \geq 0 $ using (\ref{def:ab}) and solve
\begin{align}\label{eq:yosida_cont}
P_{\epsilon,r}(u) =
\begin{cases}
\partial_tu -  \alpha K_{\eps,p}(u) +\frac{1}{r}\big( u\chi_0  +(u-1)\chi_1\big) = au -b , \qtext{in } \mathcal{Q}_T \vspace{0.2cm} \\ 
u(0,\cdot) = f \qtext{ on } \Omega
\end{cases}
\end{align}
such that $P_{\epsilon,r}(u) \to P_{\epsilon}(u)$ when $r \to 0$.

Notice that $u\chi_0 = -u^{-}$ and $(u-1)\chi_1 =(u-1)^+$, where 
we used the notation $u^+=\max(u,0)$, $u^-=-\min(u,0)$, so that $u=u^+-u^-$.  

The following result generalizes to the non-local framework the results of \cite{murea2013penalization}, establishing that the solution of \fer{eq:yosida_cont} is such that the subset of $(0,T)\times\O$ where $u(t,x)\notin [0,1]$ may be done arbitrarily small by decreasing $r$.
Thus, in the limit $r\to0$ the solution does not overpass the obstacles  $u=0$ and $u=1$.
\begin{theorem}\label{th:stability}
Let $b\in L^2(\O)$ and assume that the parameters $\alpha,\epsilon,p, r, a$ are positive. If $u\in H^{1}(0,T;L^2(\O))$ is the corresponding solution of \fer{eq:yosida_cont}, then
\begin{align*}
\int_0^T \!\!\int_\O \big(\abs{u^-}^2 + \abs{(u-1)^+}^2\big) \leq C(T)r  ,
\end{align*}
 for some constant $C(T)$ independent of $r$.
\end{theorem}
\noindent\emph{Proof. }
 Multiplying \fer{eq:yosida_cont} by $-u^-$ and integrating in $\O$, we obtain
\begin{align}
\label{proof:1}
 \frac{d}{dt}\int_\O \abs{u^-}^2 + \alpha \int_\O K_{\eps,p}(u) u^- + \frac{1}{r} \int_\O \abs{u^-}^2 = a\int_\O \abs{u^-}^2+\int_\O b u^- ,
\end{align}
where we used $\chi_1 u^- = 0$. Since $k_{\epsilon,p}$ is an odd function, the following \emph{integration by parts} formula holds
\begin{align*}
\int_\O K_{\eps,p}(u)(t,x) & u^-(t,x)dx =\\
& = \int_\O\Big(\int_\O w(x-y)k_{\epsilon ,p} (u(t,y)-u(t,x))dy\Big) u^-(t,x)dx\\
& =-\frac{1}{2}\int_\O\int_\O w(x-y)k_{\epsilon ,p} (u(t,y)-u(t,x)) (u^-(t,y)-u^-(t,x))dydx.
 \end{align*}
 Thus, in noting that $s_-$ is non-increasing as a function of $s$, we deduce 
\begin{align*}
(u(t,y)-u(t,x)) (u^-(t,y)-u^-(t,x))\leq0,
\end{align*}
and therefore, see  \eqref{ker},
\begin{align}
\label{proof:2}
 \int_\O K_{\eps,p}(u)(t,x) & u^-(t,x)dx \geq0.
 \end{align}
 Using \fer{proof:2} and the Schwarz's inequality in \fer{proof:1} we get, 
\begin{align}
\label{eq:esta1}
 \frac{d}{dt}\int_\O \abs{u^-}^2  + \frac{1}{r} \int_\O \abs{u^-}^2  \leq (a+\frac{1}{2}\big)\int_\O \abs{u^-}^2+\frac{1}{2}\int_\O b^2 .
\end{align}
Getting rid of the term  $r^{-1}\int_\O\abs{u^-}^2 \geq 0$, 
we apply Gronwall's inequality to the resulting inequality to obtain 
\begin{align*}
\int_\O \abs{u^-(t,\cdot)}^2   \leq C_1(t) \int_\O b^2 ,
\end{align*}
with $C_1(t)=t\exp((a+1/2)t)$. Using this estimate in \fer{eq:esta1} yields
\begin{align*}
 \frac{d}{dt}\int_\O \abs{u^-}^2  + \frac{1}{r} \int_\O \abs{u^-}^2 \leq C_2(t) \int_\O b^2 ,
\end{align*}
with $C_2(t) = (a+1/2)C_1(t)+1/2$. Finally, integrating in $(0,T)$ and using that $u(0,\cdot)=f \geq 0$, we obtain 
\begin{align}
\label{proof:3}
\int_0^T \int_\O \abs{u^-}^2  \leq C(T)r \int_\O b^2 ,
\end{align}
for some constant $C(T)$ independent of $r$.  
To finish the proof we must show that also 
\begin{align*}
\int_0^T \int_\O \abs{(u-1)^+}^2  \leq C(T)r  .
\end{align*}
Since, once we multiply \fer{eq:yosida_cont} by $(u-1)+$ and  integrate in $\O$, the arguments are similar to those employed to get estimate \fer{proof:3}, we omit the proof.

\section{Discretization of the limit problem $P_{\epsilon,r}(u)$}\label{sec:discretization}

In the previous section we have shown that solutions of the multivalued problem \fer{discret_model} may be approximated by the introduction of Yosida's approximants, leading to the single-valued problems $P_{\epsilon,r}(u)$ in \fer{eq:yosida_cont} that depend on the Yosida's approximation parameter $r$. We also proved that in the limit $r\to0$ the corresponding solutions lie in the relevant range of values for image processing tasks, this is, in the interval $[0,1]$.

In this section we provide a fully discrete algorithm to numerically approximate the limit solution of \fer{eq:yosida_cont} when $r\to0$.  First, we introduce a time semi-implicit Euler discretization of the evolution equation in \fer{eq:yosida_cont}, that we show to retain the stability property of its continuous counterpart, stated in Theorem~\ref{th:stability}. 

The resulting space dependent PDE is discretized by finite differences. Since the problem is nonlinear and, in addition, we want to pass to the limit $r\to0$, we introduce an iterative algorithm which renders the problem to a linear form and, at the same time,  replaces the fixed parameter $r$ by a decreasing sequence $r_j\to0$.

%
%
%
%

 
\subsection{Time discretization}

\noindent For the time discretization, let $N\in\mathbb{N}$, $\tau = T/N$, and consider the decomposition $(0,T] = \cup_{n=0}^{N-1} (t_{n},t_{n+1}]$, with $t_n = n\tau$. We denote by $u^n(x)$ to $u(t_n,x)$ and by $\chi_{i}^n(x)$ to $\chi_{i}(t_n,x)$, for $i=0,1$. 
Then, we consider a time discretization of \fer{eq:yosida_cont} in which all the terms are implicit but the diffusion term, which is semi-implicit. The resulting time semi-discretization iterative scheme is:

\vspace{0.2cm}
\noindent \textbf{Iterative Problems $P_{\epsilon,r}(u^n)$}
\vspace{0.2cm}

\noindent Given $a>0$, $b \in L^\infty (\Omega)$, $b(x) \geq 0 $, a.e. in $\Omega$ set $u^0=f$ and for $n=0,\ldots,N-1$ find $u^{n+1}:\O\to\R$ such that    
\begin{align}\label{eq:discrete_n}
P_{\epsilon,r}(u^n) = 
\begin{cases}
 u^n(x)  -\tau b(x) = 
(1-\tau a)  u^{n+1} (x)  - \tau \alpha \tilde K_{\epsilon,p}(u^n,u^{n+1})(x)\vspace{0.2cm} \\
 + \displaystyle \frac{\tau}{r}\big( u^{n+1}(x)\chi_{0}^{n+1}(x)  +(u^{n+1}(x)-1)\chi_{1}^{n+1}(x)\big), \quad \text{in } \Omega
\end{cases}
\end{align}
where, for $\tilde k_{\epsilon ,p} (s,\sigma)= \sigma\left(s^2 +\epsilon^2\right)^{(p-2)/2}$, we define
\begin{align}\label{def:kernel_approx}
\tilde K_{\epsilon,p}(u^n,u^{n+1})(x)= \int_\Omega w(x-y)\tilde k_{\epsilon ,p} (u^n(y)-u^n(x), u^{n+1}(y)-u^{n+1}(x))dy.
\end{align}
That is, only the modulus part of the diffusion term is evaluated in the previous time step. 

Equation \fer{eq:discrete_n} is still nonlinear (in fact piece-wise linear) due to the Yosida's approximants of the penalty term. In addition, its solution depends on the fixed parameter $r$ that, in view of Theorem~\ref{th:stability}, we wish to make arbitrarily small, so that the corresponding solution values are effectively constrained to the set $[0,1]$. 
To do this, we consider the following iterative algorithm to approximate the $r$-dependent solution, $u^{n+1}$, of \fer{eq:discrete_n}  \emph{when} $r\to0$.
 
\vspace{0.2cm}
\noindent \textbf{Iterative Approximating Problems $P_{j}(u^n)$}
\vspace{0.2cm}

\noindent Let $a,b$, and $f$ be as before. Let $u^0=f$ and $r_0>0$ be given. For $n=0,\ldots,N\! -\! 1$,  set  $u^{n+1}_0=u^n$. Then, for $j=0,1\ldots$, define $r_j=2^{-j}r_0$ and, until convergence, solve the following problem:  find $u^{n+1}_{j+1}:\O\to\R$ such that    
\begin{align}\label{eq:discrete_nj}
P_{j}(u^n) \!=\! 
\begin{cases}
u^n(x)  -\tau b(x) =
 (1-\tau a) u^{n+1}_{j+1} (x)  - \tau \alpha \tilde K_{\epsilon,p}(u^n,u^{n+1}_{j+1})(x) \vspace{0.2cm} \\
 + \displaystyle \frac{\tau}{r_j}\big( u^{n+1}_{j+1}(x)\chi_{0,j}^{n+1}(x)  +(u^{n+1}_{j+1}(x)-1)\chi_{1,j}^{n+1}(x)\big), \quad \text{in } \Omega
\end{cases}
\end{align}
where $\chi_{i,j}^{n+1}(x) = \chi_{\Omega\setminus\Omega_{i,j}^{n+1}}(x)$ for $i=0,1$, being 
\begin{align*}
\Omega_{0,j}^{n+1} =\{ x\in\Omega \,|\, u^{n+1}_j (x)>0\,\},\qquad \Omega_{1,j}^{n+1} =\{ x\in\Omega \,|\, u^{n+1}_j (x)<1\,\}.
\end{align*}
We use the stopping criteria 
\begin{align}
\label{def:stopping}
 \|u^{n+1}_{j+1} - u^{n+1}_{j}\|_{L^\infty(\O)} < tol,
\end{align}
for values of $tol$ chosen empirically and, when satisfied, we set $u^{n+1} = u^{n+1}_{j+1}$.

\begin{remark}	
The stability result for the time continuous problem \fer{eq:yosida_cont} stated in Theorem~\ref{th:stability} may be adapted with minor changes to the semi-implicit time discrete problem $P_{j}(u^n)$ \fer{eq:discrete_n}.
\end{remark}

\subsection{Space discretization}

For the space discretization, we consider the usual uniform mesh associated to image pixels contained in a rectangular domain, $\O=[0,L-1]\times[0,M-1]$, with mesh step size normalized to one. We denote by $x_k$ a generic node of the mesh, with $k=0,\ldots,LM-1$, and by $u[k]$ a generic function $u$ evaluated at $x_k$

To discretize the non-local diffusion term in space, we assume that $u^n$ is a constant-wise interpolator, and to fix ideas, we use the common choice of spatial kernel used in bilateral theory filtering \cite{tomasi1998bilateral}, that is, the Gaussian kernel
\begin{equation*}
w(x) = \frac{1}{C} \exp\Big(-\dfrac{\abs{x}^2}{\rho^2}\Big), 
\end{equation*}
being $C$ a normalizing constant such that $\int w = 1$. 
Assuming that the discretized version of $w$ is compactly supported in $\Omega$, with the support contained in the box $B=B_{2\rho}(x)$, we use the zero order approximation (\ref{def:kernel_approx})
\begin{align*}
\tilde K_{\epsilon,p}(u^n,u^{n+1})(x_k)\approx \sum_{m\in I_B^k} w[k,m]\tilde k_{\epsilon ,p} (u^n[m]-u^n[k], u^{n+1}[m]-u^{n+1}[k]),
\end{align*}
where $w[k,m]=w(x_k-x_m)$ and $I_B^k = \{m=0,\ldots,LM-1: \abs{x_k-x_m}< 2\rho\}$.

The values of the characteristic functions $\chi_{i,j}^{n+1}(x_k)$ of the set
$\Omega\setminus\Omega_{i,j}^{n+1}$ are the last terms of \fer{eq:discrete_nj} that we must spatially discretize. This is done by simply examining whether $u^{n+1}_j [k]>0$ or not, for $\chi_{0,j}^{n+1}[k]$, and similarly for $\chi_{1,j}^{n+1}[k]$. 

The full discretization of \fer{eq:discrete_nj} takes the form of the following linear algebraic problem:
For $k=0,\ldots,LM-1$, let $u^0[k]=f(x_k)$. For $n=0,\ldots,N-1$,  set  $u^{n+1}_0[k]=u^n[k]$. Then, for $j=0,1\ldots$  until convergence, solve the following problem:  find $u^{n+1}_{j+1}[k]\in\R$ such that    
\begin{align}
(1-& \tau a)  u^{n+1}_{j+1} [k]  - \tau \alpha \sum_{m\in I^k_B} w[k,m]\tilde k_{\epsilon ,p} (u^n[m]-u^n[k], u^{n+1}_{j+1}[m]-u^{n+1}_{j+1}[k]) \nonumber \\
& +\frac{\tau}{r_j}\big( u^{n+1}_{j+1}[k]\chi_{0,j}^{n+1}[k]  +(u^{n+1}_{j+1}[k]-1)\chi_{1,j}^{n+1}[k]\big)
 =   u^n[k]  -\tau b[k]. \label{eq:discrete_njs} 
\end{align}
The convergence of the algorithm is checked at each $j$-step according to the spatial discretization of the stopping criterium \fer{def:stopping}, that is 
\begin{align*}
\max_{0\leq k\leq LM-1} \|u^{n+1}_{j+1}[k] - u^{n+1}_{j}[k]\| < tol.
\end{align*}
When the stopping criterium is satisfied, we set $u^{n+1}[k]=u^{n+1}_{j+1}[k]$ and advance a new time step, until $n=N-1$ is reached.

\section{A simplified computational approach}\label{cinco}
In the previous sections we have deduced, through a series of approximations, a discrete algorithm to compute approximated solutions of the obstacle problem $P_{\epsilon}(u)$ in \fer{discret_model}. We have shown that our scheme is stable with respect to the approximating parameter $r$, producing solutions of problems $P_r(u)$ that, in the limit $r\to0$, lie effectively in the image value range $[0,1]$, apart from producing the required edge preserving saliency detection on images.

In this section, by introducing some hard nonlinearities (truncations) to replace one of the iterative loops of \fer{eq:discrete_njs}, we provide a simplified algorithm for solving a problem closely related to \fer{discret_model}. In addition, we use an approximation technique, based on the discretization of the image range, to compute the non-local diffusion term. These modifications allow for a fast computation of what we demonstrate to be fair approximations to the solutions of the original problem \fer{discret_model}.

Considering the time discrete problem \fer{eq:discrete_n}, we introduce two changes which greatly alleviate the computational burden:
\begin{enumerate}
 \item Compute the non-local diffusion term fully explicitly, and 
 \item Replace the obstacle term by a hard truncation.
 \end{enumerate}
Thus, we replace problem $P_{\epsilon,r}(u^n)$ in \fer{eq:discrete_n} by the following which can be deduced from problem $P_\epsilon(u)$ in (\ref{discret_model}) using the two above strategies.

\vspace{0.2cm}
\noindent \textbf{Explicit Truncated Problems $P_{\epsilon,0}(u^n)$} 
\vspace{0.2cm}

\noindent Given $u^0=f$, and for $n=0,\ldots,N-1$, find $u^{n+1}:\O\to\R$ such that    
\begin{align}\label{eq:discrete:simple_n}
P_{\epsilon,0}(u^n) = 
\begin{cases}
(1-\tau a) u^{n+1} (x)  = \tau \alpha K_{\epsilon,p}(u^n)(x) +   u^n(x)  -\tau b(x) \text{ in } \Omega
\end{cases}
\end{align}
followed by a truncation of $u^{n+1}$ within the range $[0,1]$. 
Observe that, as remarked in \cite{perez2011numerical}, the explicit Euler scheme is well suited for non-local diffusion since it does not need a restrictive stability constraint for the time step, as it happens for the corresponding local diffusion. This is related to the lack of regularizing effect in non-local problems.  
Spatial discretization of \fer{eq:discrete:simple_n}  leads to the following algorithm which we shall refer as the {\em patch based scheme}: 

Set $u^0=f$. For $n=0,\ldots,N-1$, and for $k=0,\ldots,LM-1$, compute 
\begin{equation}\label{ec.patch}
  u^{n+1} [k]  = \left( \frac{\tau \alpha }{1- \tau a} \right) \sum_{m\in I^k_B} w[k,m]k_{\epsilon ,p} (u^n[m]-u^n[k]) +  u^n[k]  -\tau b[k] 
\end{equation}
and truncate $u^{n+1}[k]$
\begin{align*}
\tilde u^{n+1}[k]=\min(1,\max(0,\tilde u^{n+1}[k])),
\end{align*}

We shall show that there are very small differences between the solutions of the explicit truncated problem $P_{\epsilon,0}(u^n)$ and the solutions of the $P_{\epsilon,r}(u^n)$ problems for $r$ sufficiently small. Nevertheless, the numerical scheme is greatly improved and much more efficient because costly iteration in $r$-loop is avoid as it can be seen in section \fer{sec:experiments}.
We finally describe the efficient approach of \cite{yang2009real} (see also \cite{galiano2015fast},   and and \cite{galiano2016well} for a related approach) that we use for computing the sum in \fer{ec.patch},            
corresponding to the non-local diffusion term, by discretizing also the range of image values. Let $q = \{q_1,\ldots,q_Q\}$ be a quantization partition,  with 
$0=q_1<q_2<\ldots<q_{Q-1}<q_Q =1$, where $Q$ is the number of quantization levels. 
Let $v:\O\to[0,1]$ be a quantized function, that is, taking values on $q$. 
For each $i = 1,\ldots,Q$, we introduce the discrete convolution operator 

\begin{equation}\label{op}
 K_{\epsilon,p}^{i}(v) [k]=\sum_{m\in I^k_B} w[k,m] k_{\epsilon ,p} (v[m]-q_i) ,
\end{equation}
 where we recall that $w[k,m]=w(x_k-x_m)$. We then have 
 \begin{align*}
 K_{\epsilon,p} (v)[k] =K_{\epsilon,p}^i (v)[k] \qtext{if }v[k] = q_i, \qtext{for some }i=1,\ldots,Q.
 \end{align*}
 Notice that, for any $v$ taking values in $q$, the computation of each $K_{\epsilon,p}^i (v)$ may be carried out in parallel by means of fast convolution algorithms, e.g. the fast Fourier transform.

It is possible that, after a time iteration, a quantized iterand $u^n$ leads to values  of $u^{n+1}$ not contained in the quantized partition $q$, implying that the new operators $K_{\epsilon,p}^i (u^{n+1})$ should be computed in a new quantization partition, say $q^{n+1}$. Since, for small time step, we expect $q^n$ and $q^{n+1}$ to be close to each other, we overcome this inconvenient by rounding $u^{n+1}$ to the closest value of the initial quantization vector, $q$, so that this vector remains fixed.

The final simplified algorithm which we call the  {\em kernel based scheme}  is then: 

Set $u^0=f$. For $n=0,\ldots,N-1$, and for each $k=0,\ldots,LM-1$, perform the following steps:

\begin{itemize}
\item \textbf{Step 1. } If $u^n[k]=q_i$ then
using \fer{op}

\begin{equation}\label{ec.kernel}
\tilde u^{n+1} [k]  = \frac{1}{1- \tau a}\Big(\tau \alpha 
K_{\epsilon,p}^{i}(u^n) [k] 
+  q_i  -\tau b[k]\Big). 
\end{equation}

\item \textbf{Step 2. } $u^{n+1} [k] = q_j$, where  $j = \underset{1\leq i\leq Q}{\text{argmin}}\abs{q_i-\tilde u^{n+1} [k]}$.
\end{itemize}

\newpage

\section{Numerical experiments}\label{sec:experiments}

In this section we describe the experiments that support our conclusions. First, we compare the use of hard truncation (in the fully explicit numerical scheme \fer{eq:discrete:simple_n}) with the iterative scheme ($r\to 0$) when the Yosida's Approximants are used to solve the discrete problem \fer{eq:discrete_nj}. As a second test, the proposed kernel based approach in \fer{ec.kernel} is compared with the patch based scheme in \fer{ec.patch}.  
As an application of the above schemes, we test our approach over the BRATS2015 dataset \cite{menze2015multimodal}. Finally, we show that our model can be generalized presenting some preliminary  results which extend our variational non-local saliency model to 3D volumes.

\subsection{Experiment 1: Comparison between limit approximation  and truncation }
In this experiment we show the differences between the limit approximation $r \to 0$ described in section \ref{sec:discretization} and the proposed truncation alternative. We recall that the purpose of such hard truncation is to get rid of the $r$-loop in the numerical resolution, boosting the computation efficiency. In practice, instead of using the stopping criteria (\ref{def:stopping}), is sufficient (and more efficient) to fix a small number of iterations which results into 5 in the $r$-loop starting with $r^1=0.5$ and setting $r^{j+1} = 2^{-j}r^{j},~~j = 1,\ldots,5$. We found in our experiments that this is enough in order to ensure that the final output of the approximating scheme (\ref{eq:discrete_nj}) is very close to the solution of \fer{ec.patch}.

Each $j$-step consists of solving the equation (\ref{eq:discrete_nj}) which is carried out through a conjugate gradient descent algorithm. This is an inner loop for each $j$-step which increases substantially the global time execution of the algorithm. In order to show that the hard truncation is a good strategy to get rid of the $r$-loop we compute the relative differences 
$||u_J - u_T||_2 /||u_T||_2$  between each $j$-step image ($u_J$) and the truncated version ($u_T$) of the $n$-step solution.

\begin{figure}[H]
  \includegraphics[width=\textwidth]{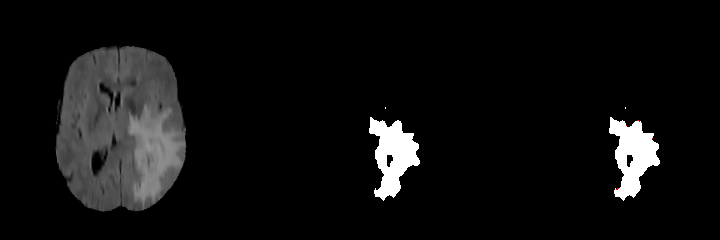}
\caption{From left to right: input image, $r$-convergence approximation and hard truncation approximation. Differences are colored in red.}
\label{fig:comparison_1}       
\end{figure}
At each step of the $r$-loop, the relative difference from the final truncated version is reduced. Figure \ref{fig:comparison_1} depicts the qualitative difference of using truncation. For all the subjects we tested the results clearly show that the same saliency (tumor) region is detected in both images. The differences, barely visibles, are colored in red. Indeed only few pixels differ from the assumed correct computation through the $r$-convergence, which justifies the use of the hard truncation.
\begin{figure}[H]
\centering
  \includegraphics[width=1\textwidth]{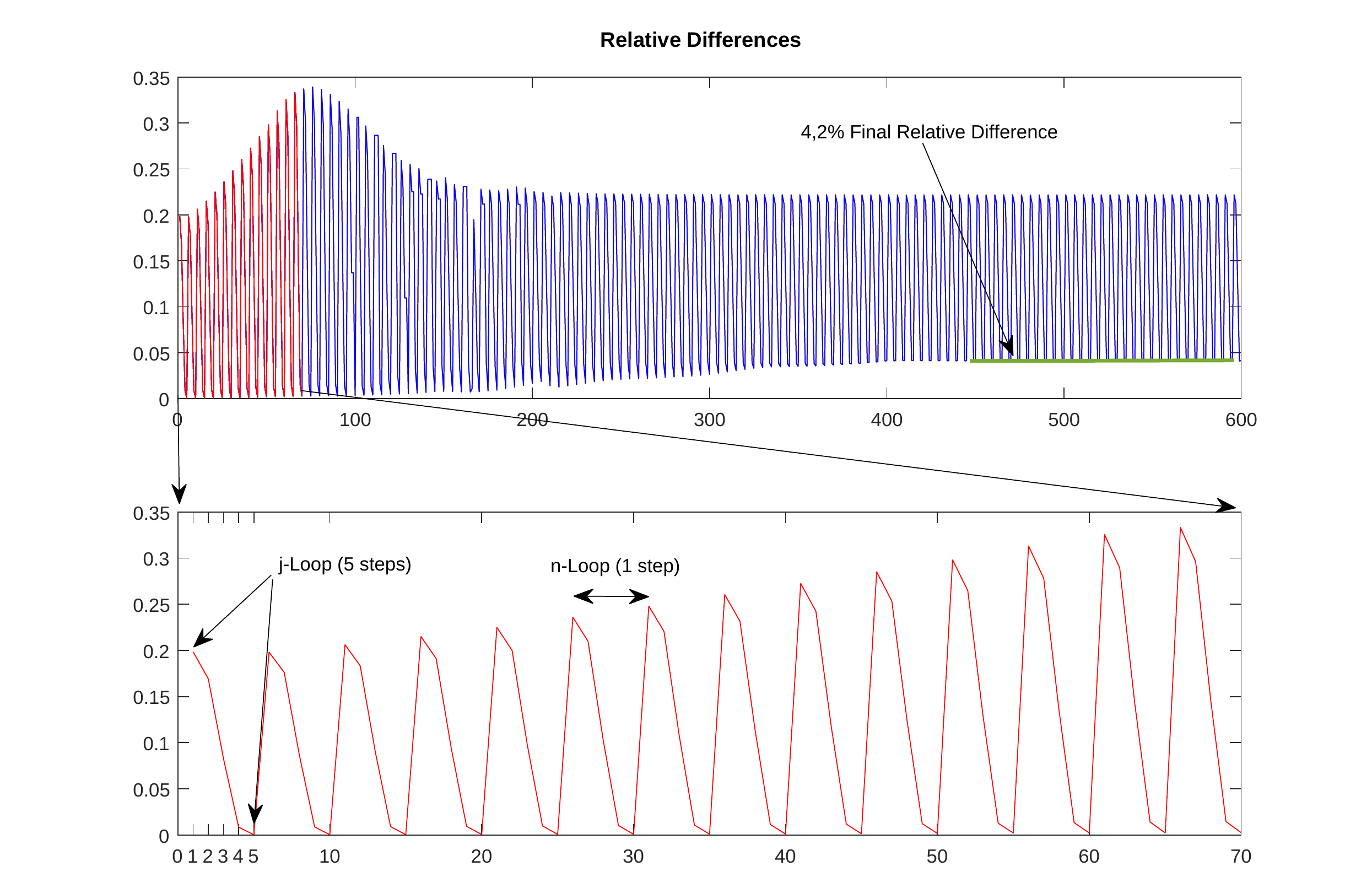}
\caption{Relative differences: $r-$convergence vs hard truncation. Stabilization of the $r \to 0$ limit. The outer time $n$-loop, $n=1$ ... N is considered together with the inner $j$-loop, $j=1$ ... J. The jumps from $u_5$ to $u_6$, $u_{10}$ to $u_{11}$ etc are caused by the truncation with J fixed to 5 for each n. We see in Figure \ref{fig:comparison_1} that the (binary) output solution is practically indistinguishable from the almost binary approximations in r. The final outputs only differ a 4,2\%, which in practice turns out to be few single pixels.}
\label{fig:comparison_2}       
\end{figure}
\subsection{Experiment 2: Kernel based approach}

In this experiment we compare the time execution between patch based numerical resolution and the proposed kernel approach based on \cite{yang2009real}. Taking advantage of the fact that convolutions can be fast computed in Fourier domain, we use a GPU implementation to carry out these experiments. In both cases we fix the same hyper-parameters and perform a sweep where $\rho = {2,3,\ldots,30}$ is the kernel radio and $q = {2^3,2^4,\ldots,2^{11}}$ are the quantization levels. The tests are performed over 4 brains (2 slices per brain) and results are averaged.

\begin{figure}[H]
  \centering
   \includegraphics[width=0.8\textwidth]{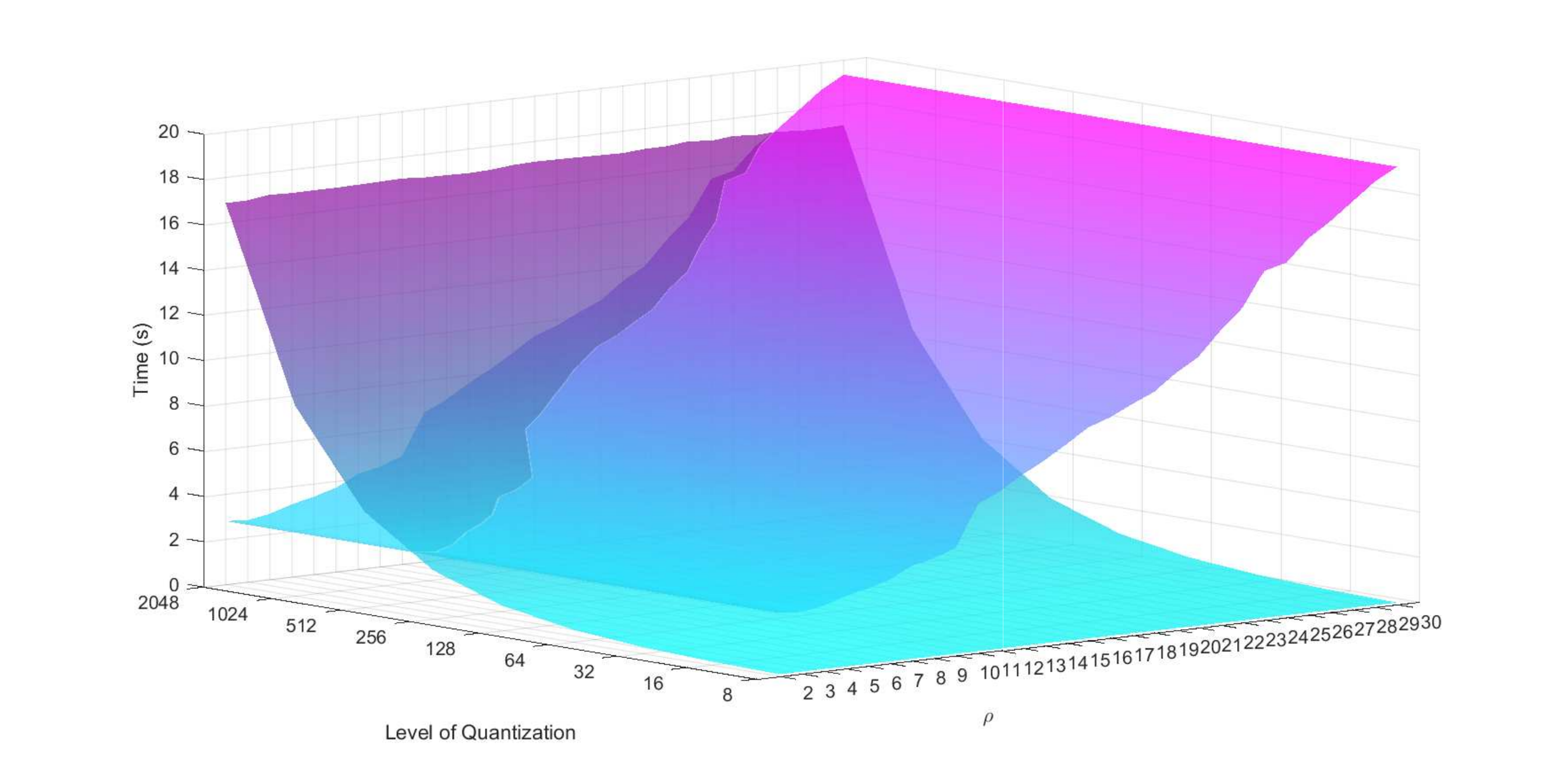}
  \caption{Time comparison between patch based and kernel based resolution. The surface which remains constant with the quantization levels corresponds to the patch based resolution while the other one corresponds to the kernel based resolution. Using a kernel based approach allows a nearly invariant dependency w.r.t. the size of $\rho$ (radius of the kernel), which in turn promotes the non-locality effect.}
  \label{fig:comparison_time}
 \end{figure}

Notice that in a classical patch based approach no quantization is required and the time execution will grow up only with the size of the considered region ($B_{2\rho}(x)$). On the contrary, a kernel based resolution remains robust to different kernel sizes while the time execution relays mainly in the number of quantization levels as it can be seen in figure \ref{fig:comparison_time}. This justifies the use of the kernel based method whereas it allows to use bigger kernels so properly modelling the non-local diffusion term. 

\subsection{Experiment 3: MRI Dataset}
\begin{figure}[H]
\centering
 \includegraphics[width=0.85\textwidth]{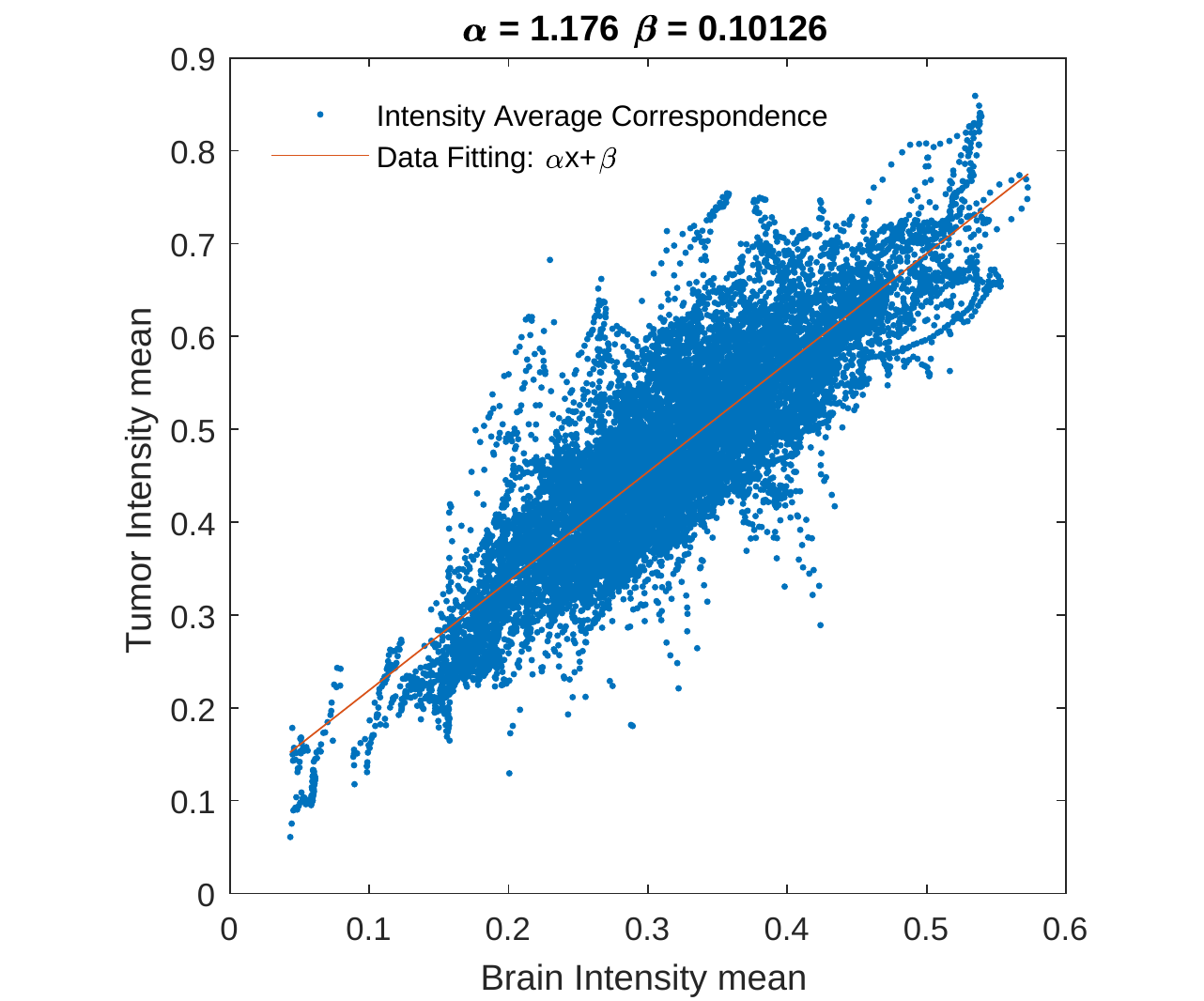}
  \caption{Linear regression for data: {$\mu_{brain}$, $\mu_{tumor}$}}
  \label{fig:fitting}
\end{figure}

We then apply our above findings testing the whole BRATS2015 dataset \cite{menze2015multimodal}. In order to highlight the importance of the proposed non-local non-convex hyper-Laplacian operators we shall consider the behavior of the model when only the $p$-parameter is modified. For such purpose we need  to introduce an automated rule for the $\delta$-parameter estimation to avoid a manual tuning of the model for each image of the dataset. This will result in a sub-optimal performance of the model in terms of accuracy. Nevertheless, it will provide us a good baseline of how good is our proposed model. Observing that $1/\delta$ acts as threshold between classes (background and foreground), we seek a rule to determine such threshold for each image leading to an approximatively correct estimation of $\delta$. By averaging the whole given brain (values of pixels where there is brain), and comparing with the average of the tumor intensities, it turns out (see Figure \ref{fig:fitting}) that the relationship is pretty linear, and a simple lineal regression gives a prediction of the mean of the tumor in the considered image:
$$
\mu_{tumor} \approx \alpha \mu_{brain} + \beta = 1.176 \mu_{brain} + 0.101
$$
We then select a reference threshold. A simple choice is to compute the average between $\mu_{brain}$ and $\mu_{tumor}$ in form
$(1/\delta)=(\mu_{brain} + \mu_{tumor})/2$.
Finally, since it is always possible to compute the average of the whole brain ($\mu_{brain}$), we end up with the following rule for the $\delta$-parameter estimation (depending on the specific image considered):

$$
\delta(\mu_{brain}) = \frac{2}{(1+\alpha)\mu_{brain} + \beta}
$$

Our proposed model includes hyper-Laplacian non-local diffusion terms by setting $p<1$. We also compare different values of $p = {2,1,0.5}$ with the same parametrization and show in table \ref{tab:ps}, where typical reference metrics (\cite{powers2011evaluation}) are reported, that the DICE metric is monotonically increased as $p$ is decreased. 

\begin{table}[h]
\centering
\caption{Results for different $p$ values over the whole BRATS2015 dataset.}
\label{tab:ps}
\begin{tabular}{@{}llll@{}}
\toprule
        & Precision & Recall  & DICE    \\ \midrule
Naive Threshold    & 0.4431   & 0.7904 & 0.5299 \\
$p=2$   & 0.5959   & 0.\textbf{7904} & 0.6484 \\
$p=1$   & 0.6799   & 0.7798 & 0.7013 \\
$p=0.5$ & \textbf{0.7658}   & 0.7321 & \textbf{0.7276} \\ \bottomrule
\end{tabular}
\end{table}

\subsection{Experiment 4: 3D versus 2D model}
\begin{figure}[H]
\centering
  \includegraphics[width=1\textwidth]{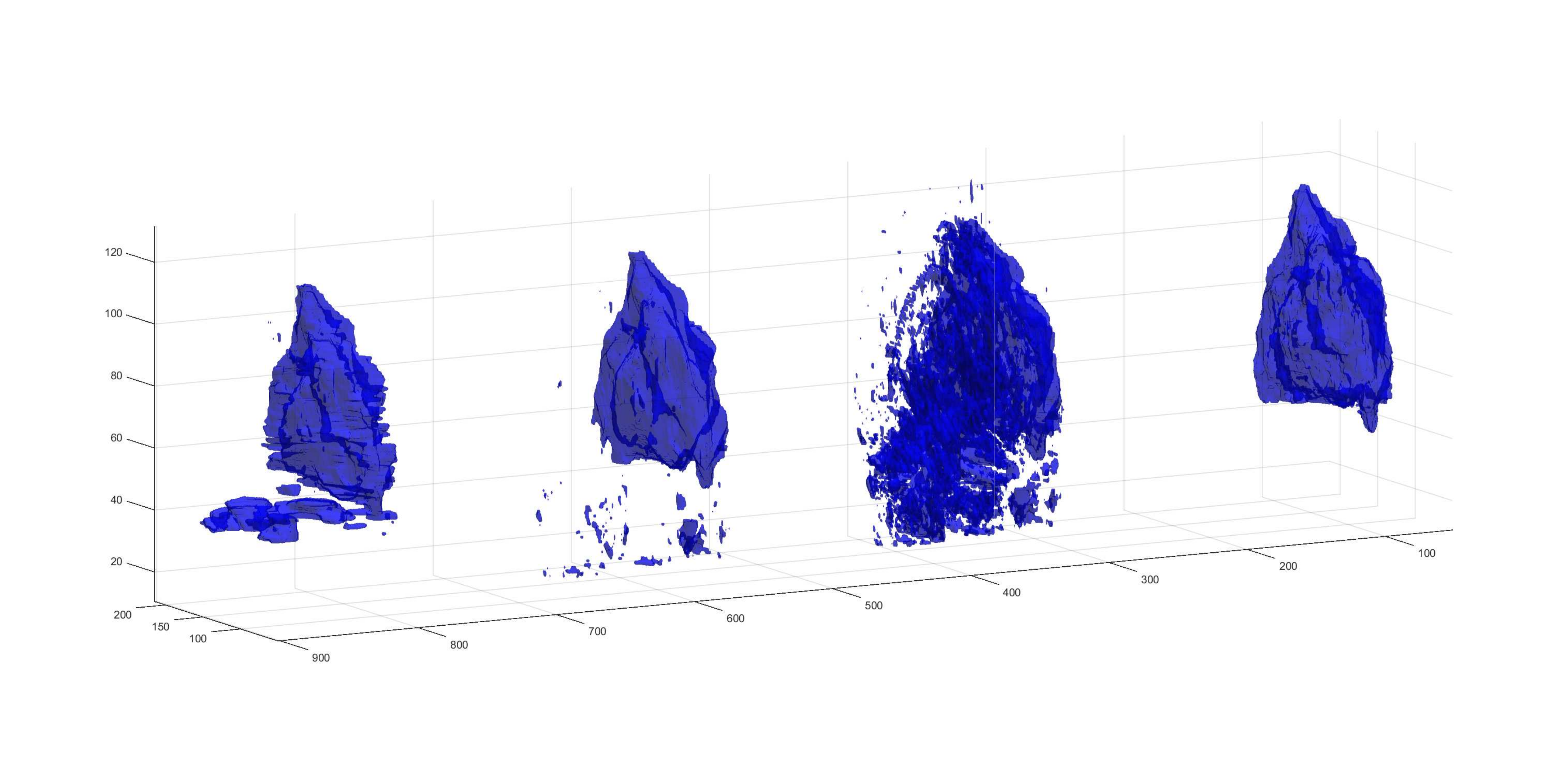}
\caption{From left to right: 2D reconstruction (our proposed model applied per slice), 3D reconstruction (our proposed model applied to the whole volume), Naive threshold, Ground-truth}
\label{fig:3d}       
\end{figure}

Focusing in the particular application of brain tumor segmentation, it is reasonable to think that processing each image (slice) independently will result in a sub-optimal output since no axial information is taken into account. Our model can be easily extended to process 3D brains volumes and not only 2D images (gray-scale images). In such a way, the non-local regularization will prevent from errors using 3D spatial information. The results are greatly improved  as it is reported in table \ref{tab:3dvs2d} and shown in figure \ref{fig:3d}. It is easy to see that some artifacts arise when processing the volume per slice (2D , first image), which disappear if a 3D processing is considered (second image). False positives can also be avoided in this 3D approach obtaining a cleaner image that results in a very high accuracy in common metrics (Precision, Recall and DICE, see table \ref{tab:3dvs2d}).
However, even thought the results show promising performance using this 3D scheme in comparison to a 2D processing so improving the final segmentation, the real challenge consists of choosing a correct and robust $\delta$-parameter feasible for the whole volume. This is due to the non-homogeneous contrast and illumination in different regions of the MRI image which depends on the acquisition step. In future work, we will introduce some recent advances in Deep-Learning parameter estimation to obtain the optimal model parametrization which can enhance further the performance of our model (see \cite{ramirez2018optimization}).
\vspace{0.1cm}
\begin{table}[h]
\centering
\caption{Results of a 3D and 2D resolution with the same parametrization}
\label{tab:3dvs2d}
\begin{tabular}{@{}llll@{}}
\toprule
 & Precision & Recall & DICE  \\ \midrule
Naive Threshold & 0.5321 & 0.8008 & 0.6393 \\ 
2D   & 0.8658 & 0.7986 & 0.8308 \\
3D   & \textbf{0.9649}  & \textbf{0.8656}  & \textbf{0.9125}  \\  \bottomrule
\end{tabular}
\end{table}
\newpage
\section{Conclusions}
\label{sec:conclusion}
Dealing with recent challenging problems such as automatic saliency detection of videos or fixed 2D images we propose a new approach based on reactive flows which facilitates the saliency detection task promoting binary solutions which encapsulate the underlying classification problem. This is performed in a non-local framework where a bilateral filter is designed using the NLTV operator. The resulting model is numerically compared with a family of non-smooth non-local non-convex hyper-Laplacian operators. The computational cost of the model problem solving algorithm is greatly alleviated by using recent ideas on quantized convolutional operators filters making the approach practical and efficient. Numerical computations on real data sets in the modality of MRI-Flair glioblastoma automatic detection show the performance of the method.

\noindent In this work we have presented a new non-local non-convex diffusion model for saliency detection and classification which has shown to be able to perform a fast foreground detection when it is applied to a FLAIR given image. The results reveal that this method can achieve very high accurate statistics metrics over the ground-truth BRATS2015 data-set \cite{menze2015multimodal}. Also, as a by-product of the reactive model, the solution has, after few iterations, a reduced number of quantized values making simpler the final thresholding step.
Such a technique could be improved computationally by observing that the diffusion process combined with the saliency term evolves producing more cartoon like piece-wise constant solutions which can be coded with less number of quantization values while converging to a binary mask. This is related to the absorption-reaction balance in the PDE where absorption is active where the solution is small, $u\approx 0$ and the reaction is active where $u\approx 1$. 
The non-local diffusion properties of the model also allow to detect salient objects which are not spatially close as well as connected regions (disjoint areas). This can be useful in many other medical images modalities, specially in functional MRI (fMRI). Non-convex properties, meanwhile, promote {\em sparse} non-local gradient, pushing the solution to a cartoon piece-wise constant image. Both characteristics combined with our proposed concave energy term results in a promising accurate and fast technique suitable to be applied to FLAIR images and others MRI modalities.

\section*{\uppercase{Acknowledgments}}

The first and the last authors' research has been partially supported by the Spanish Government research funding ref. MINECO/FEDER TIN2015-69542-C2-1 and the Banco de Santander and Universidad Rey Juan Carlos Funding Program for Excellence Research Groups ref. “Computer Vision and Image Processing
(CVIP)”. The second author has been supported by the Spanish MCI Project MTM2017-87162-P.

\newpage
\bibliographystyle{spmpsci}
\bibliography{biblio}

\end{document}